\documentclass[conference]{IEEEtran}
\IEEEoverridecommandlockouts
\usepackage{cite}
\usepackage{url}
\usepackage{amsmath,amssymb,amsfonts}
\usepackage{algorithmic}
\usepackage{graphicx}
\usepackage{textcomp}
\usepackage{xcolor}
\usepackage{cite}

\usepackage{comment}
\def\BibTeX{{\rm B\kern-.05em{\sc i\kern-.025em b}\kern-.08em
    T\kern-.1667em\lower.7ex\hbox{E}\kern-.125emX}}

\ifCLASSOPTIONcompsoc
 \usepackage[caption=false,font=normalsize,labelfont=sf,textfont=sf]{subfig}
\else
 \usepackage[caption=false,font=footnotesize]{subfig}
\fi
\usepackage{amsmath}
\usepackage{graphicx}
\usepackage{xcolor}
\def\BibTeX{{\rm B\kern-.05em{\sc i\kern-.025em b}\kern-.08em
    T\kern-.1667em\lower.7ex\hbox{E}\kern-.125emX}}
\usepackage{hyperref}

\hypersetup{
    colorlinks = true,
    linkcolor  = black,
    filecolor = black,      
    urlcolor = black,
    citecolor=black,
} 

\makeatletter
 \let\old@ps@headings\ps@headings
 \let\old@ps@IEEEtitlepagestyle\ps@IEEEtitlepagestyle
 \def\confheader#1{%
 \def\ps@IEEEtitlepagestyle{%
 \old@ps@IEEEtitlepagestyle%
 \def\@oddhead{\strut\hfill#1\hfill\strut}%
 \def\@evenhead{\strut\hfill#1\hfill\strut}%
 }%
 \ps@headings%
 }
 \makeatother

\confheader{%
2021 International Conference on Automation, Control and Mechatronics for Industry 4.0 (ACMI), 8-9 July 2021, Rajshahi, Bangladesh
\hspace{3.6 in}}

 \usepackage[pscoord]{eso-pic}
\newcommand{\placetextbox}[3]{
 \setbox0=\hbox{#3}
 \AddToShipoutPictureFG*{ \put(\LenToUnit{#1\paperwidth},\LenToUnit{#2\paperheight}){\vtop{{\null}\makebox[0pt][c]{#3}}}
 }
 }
 \placetextbox{.21}{0.055}{\small{978-1-6654-3843-8/21/\$31.00 ©2021 IEEE}}

\begin{document}

\title{Demand Forecasting in Smart Grid Using
Long Short-Term  Memory}

\author{
    \IEEEauthorblockN{\textcolor{black}{.}}
     \hfill
    \and
    \IEEEauthorblockN{  Koushik Roy$^1$, Abtahi Ishmam$^1$ and Kazi Abu Taher$^2$
    \\ 
    Department of EECE$^1$ \& Department of ICT$^2$\\ 
    Military Institute of Science \& Technology (MIST)$^1$, Dhaka - 1216, Bangladesh\\ 
    Bangladesh University of Professionals (BUP)$^2$, Dhaka - 1216, Bangladesh}\\ 
    \small{ \texttt{
     rkoushikroy2@gmail.com; abtahiishmam3@gmail.com; kataher@gmail.com;} }
    \hfill
    \and
    \IEEEauthorblockN{\textcolor{black}{.}}
}

\maketitle

\begin{abstract}
Demand forecasting in power sector has become an important part of modern demand management and response systems with the rise of smart metering enabled grids. Long Short-Term Memory (LSTM) shows promising results in predicting time series data which can also be applied to power load demand in smart grids. In this paper, an LSTM based model using neural network architecture is proposed to forecast power demand. The model is trained with hourly energy and power usage data of four years from a smart grid. After training and prediction, the accuracy of the model is compared against the traditional statistical time series analysis algorithms, such as Auto-Regressive (AR), to determine the efficiency. The mean absolute percentile error is found to be 1.22 in the proposed LSTM model, which is the lowest among the other models. From the findings, it is clear that the inclusion of neural network in predicting power demand reduces the error of prediction significantly. Thus, application of LSTM can enable more efficient demand response system.\\ 
\end{abstract}

\section{Introduction}
With the advancement of Internet of Things (IOT), the smart grids have gained the ability of real time metering termed as smart metering. This intelligent metering technology greatly benefits power demand forecasting because of its requirement of constant information flow between the smart grid and the consumer. The availability of past energy usage data makes it possible to test new approaches and search for optimizations in the existing forecasting models \cite{reka2018future}.

Power companies and power consumers both need power demand forecasting for an uninterrupted workflow. Forecasting provides some benefits such as real time pricing and irregularity detection. It has different application for individual data vs aggregated data \cite{hock2020using}. All consumer load data, in a particular area, is combined in order to get the aggregated forecasting result. This forecasting is used by the power company to predict the overall power demand trend and take necessary steps beforehand so that no load shedding occurs. On the contrary the individual data assists in other ways such as abnormal meter measurements or unexpected failure in meter technology. Also it can assist in detecting deliberate manipulation in power utilities. The forecasting is such an important part in power industry as it emphasises on the proper resource management and optimization thus making the smart grid more efficient and self sustaining\cite{siano2014demand}.

Time-series modeling is a very popular statistical model. On one hand there are some traditional statistics based models for time series analysis such as ARMA \cite{gross1987short} \cite{mashima2012evaluating}, ARIMA \cite{alberg2018short} \cite{cho1995customer}. Hong et al. endorsed
multiple linear regression for the modeling of hourly energy demand using seasonality \cite{hong2010modeling}. In 2019, D. Rolnick et al.  presented that energy forecasting is recognized as one
of the most significant contribution areas of Machine Learning(ML) toward transitioning to a electrical infrastructure \cite{rolnick2019tackling}. Existing investigation into the effectiveness of neural network based models such as Artificial Neural Networks provide good accuracy in demand forecasting compared to traditional models \cite{al1999artificial}. Over a long forecasting horizon, LSTM-based RNN models can accurately anticipate complicated nonlinear, non-stationary, and nonseasonal uni-variate electric load time series. Deep Neural Networks (DNNs) and traditional RNNs can not learn temporal sequences or long-term dependencies as well as LSTM-RNN architectures can \cite{zheng2017electric}. In addition, Cheng et al. presented that PowerLSTM, which is the first power demand forecasting model based on LSTM, incorporates time series features, weather features, and calendar features  \cite{cheng2017powerlstm}. It can outperform some models adopted in recent research works such as,  Support Vector Regression (SVR) and Gradient Boosting Tree (GBT).

So, it is evident from existing research that application of RNN models such as LSTM may provide much better results with higher accuracy and consistency than statistical models.
In this paper, an LSTM model is proposed to predict the power demand of smart grids and to calculate the error rate. The results prove to be promising as the model greatly outperforms traditional statistical models.

The paper has been organized in four sections. In Section I, an overview of this
research work has been presented with relevant literature review. In Section II, 
the proposed system model for forecasting power demand is discussed. In Section III, the performance of the proposed model has been
analyzed and compared against different models such as Autoregressive. Finally, the paper was concluded with Section IV.
\section{System Model}
\begin{figure}[h]
\centerline{\includegraphics[width=.85\linewidth,height=12cm]{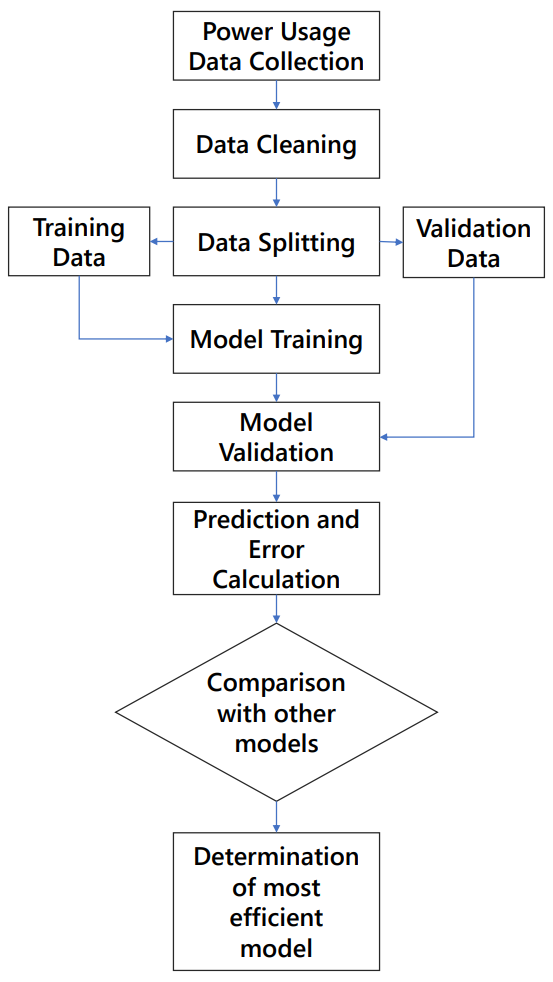}}
\caption{Detailed flow diagram illustrating the methodology}
\label{flow chart}
\end{figure}

The demand prediction model used in this paper is based on LSTM where the parameters were tuned for higher accuracy. In order to get accurate predictions from the model, an extensive amount of smart grid power usage data is necessary. So, smart grid power usage data from a smart grid in Spain is used. Then this data is divided into two sections: the training data set and the validation dataset. For the model, the training dataset is used. After that, all the models were tested with the validation dataset to ensure the models were performing as expected. After the validation step, predictions were made using both machine learning and statistical models.
In order to establish a valid comparison between all the models, MAE (Mean Absolute Error) is calculated for all predictions. A detailed flow diagram illustrating the methodology containing every major section is given in Fig. \ref{flow chart}.

\subsection{Data Source}
The dataset has electrical generation, consumption, pricing, and weather data collected in the period of 4 years for Spain. A public portal for Transmission Service Operator (TSO) is used to retrieve the consumption and generation data. The dataset is unique in that it includes hourly data for electrical usage as well as the TSO's forecasts for consumption and pricing. This allows future predictions to be compared to existing state-of-the-art forecasts in use in the industry \cite{url1}.

\subsection{Data Exploration}
Before building any neural network model, the dataset needs to be explored in several ways to gather insight about the data so that it can be used properly to optimize the output. In the energy dataset, there are 29 columns in total. They consist of generated energy of different sources and load power demand. 

To ensure better forecasting, only the columns containing useful information (Total load forecast and Total load actual) were extracted from the entire energy dataset. The Table \ref{datasetlabel} contains the useful information about those specific columns in brief. In actual load data there are 35k data points in total, where 36 missing values, mean is 28.7k, standard deviation is 4.57k, minimum value 18k and maximum value 41k. From the represented data in the table, it is evident that the default load forecasting method used in the industry is almost on point. Before passing the data into the model the missing values are dropped to avoid any miscalculation.

\begin{table}[h]
\caption{Total load forecast and total load actual dataset containing useful information}
\begin{center}
\begin{tabular}{|c|c|c|}
\hline
\textbf{Feature} & \textbf{Total load forecast}& \textbf{Total load actual}\\ \hline

Valid               &            35.1k & 35.0k\\ \hline
Mismatched               &           0 & 0\\ \hline
Missing               &           0 & 36\\ \hline
Mean      &          28.7k & 28.7k\\ \hline
Standard Deviation                      &4.59k & 4.57k\\ \hline
Quantities (Min)      &          18.1k & 18k\\ \hline
Quantities (Max)      &          41.4k & 41k\\ \hline

\end{tabular}
\label{datasetlabel}
\end{center}
\end{table}

\begin{figure}[h]
\centerline{\includegraphics[width=1\linewidth, height=6cm]{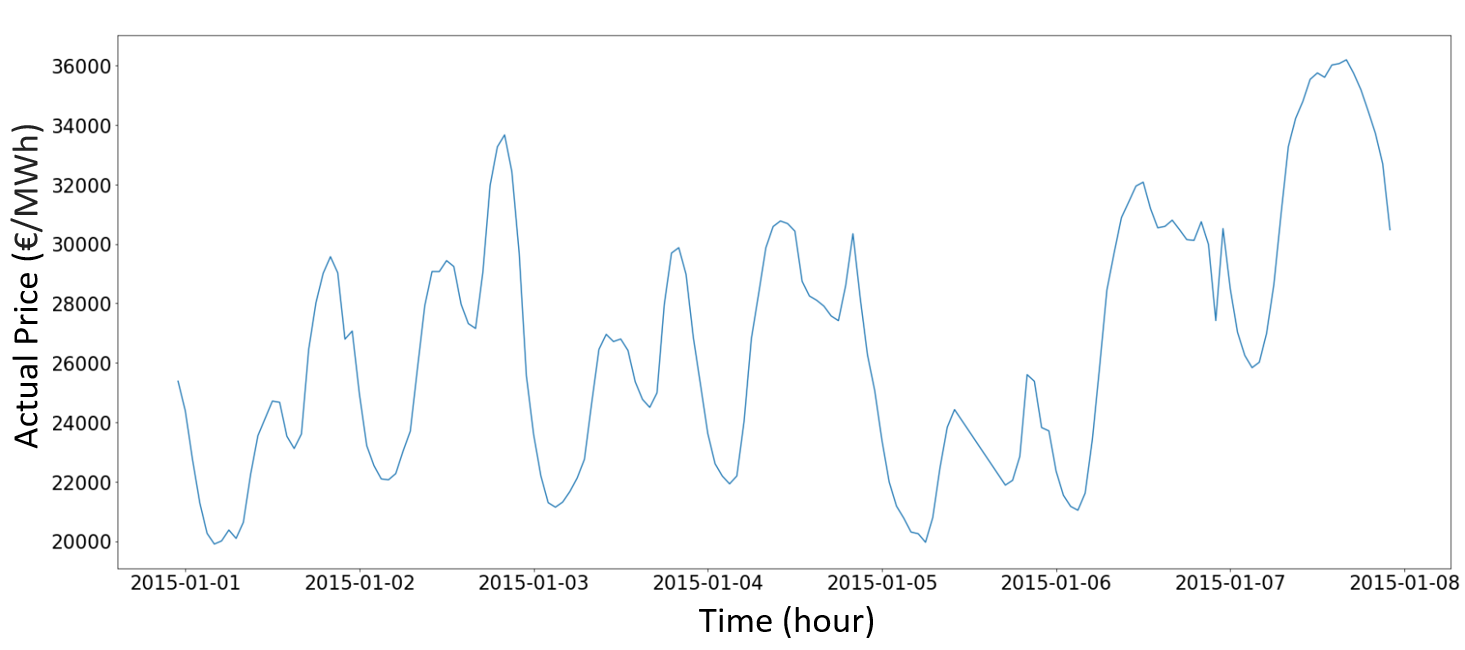}}
\caption{The figure representing actual total load in MWh collected in every hour for one week}
\label{Actual Total Load for first 4 weeks}
\end{figure}

In the dataset, hourly electrical data can be found and the data has been explored in the hope of finding useful correlation among various variables. From the Fig. \ref{Actual Total Load for first 4 weeks} the actual hourly load data can be considered as time series data.

\subsection{Correlation}
In time series analysis and forecasting, autocorrelation and partial autocorrelation plots are commonly used. These are graphs that demonstrate the strength of a relationship between an observation in a time series and previous observations.

In Table \ref{tab111} the correlation between the total load actual and the rest of the major features has been presented. All major features give positive correlation thus make them viable for predicting the load demand. The forecasted load value gives a positive correlation of 0.99 thus indicating that the selected parameters are suitable for the forecast.

\begin{table}[h]
\caption{Correlations between the total load actual and the rest of the major features}
\begin{center}
\begin{tabular}{|c|c|}
\hline
\textbf{Feature} & \textbf{Correlation Coefficient}\\ \hline
total load forecast     &        0.9951 \\ \hline
generation fossil gas               &            0.5489\\ \hline
generation fossil oil                &           0.4971\\ \hline
generation hydro water reservoir      &          0.4795\\ \hline
price day ahead                       &          0.4739\\ \hline
price actual                           &         0.4361\\ \hline
forecast solar day ahead                &        0.4044\\ \hline
generation fossil hard coal            &         0.3966\\ \hline
generation solar                        &        0.3962\\ \hline
generation fossil brown coal/lignite     &       0.2805\\ \hline
generation other renewable              &        0.1817\\ \hline
generation other                        &        0.1007\\ \hline
generation nuclear                    &          0.0857\\ \hline
generation biomass                   &           0.0833\\ \hline
generation waste                     &           0.0773\\ \hline
generation wind onshore              &           0.0401\\ \hline
forecast wind onshore day ahead      &           0.0376\\ \hline

\end{tabular}
\label{tab111}
\end{center}
\end{table}

The Pearson correlation coefficient is measured in statistics by a linear correlation between the two data sets. The covariance of two variables divided by the standard variables. Therefore the results always have a value between $-$1 and 1 are essentially a standard covariance measurement. The measure can only mirror a linear correlation of variables and ignore many others like covariance itself. While not much of the matrix can be achieved, many of the features are highly interrelated.

\subsection{Data Preprocessing and Windowing}
Data preprocessing is the process by which missing (null) values are cleaned, normalized, filled out. 36 missing values were found in the used dataset. In the preprocessing step, the value of look-back was kept 25. The data was restructured to make it compatible for input into the LSTM model. Each input data was a list of 25 hours of power consumption and the output data for that particular input was the power consumption for the next hour. The dataset was split into train and test set where 80\% of the total data was used for model training and the remaining 20\% was used for testing. 

\subsection{Model Description}
In this research work, a model based on LSTM architecture is proposed and the suitability of the model is validated by comparing with other models such as AR, MA, ARMA and ARIMA. Firstly, the AR model anticipates future conduct based on past conduct. The process essentially involves a linear regression of the current series data against one or more preceding values of the same series. Secondly, MA is a simple technical analysis tool, which allows price data to be smooth by constantly updating the average price. The average time frame may be 10 days, 20 days, 30 weeks or any other time frame. In this case, an average of 24 hours of moving average is taken since it is also appropriate for LSTM. Thirdly, An ARMA, or autoregressive moved average, is used to describe two polynomials as weakly stationary stochastic times. The first is for self-regression and the second for the moving average. Fourthly, ARIMA describes a given time series on the basis former values of its own, i.e. its own deficits and lagged prediction errors to predict future values. The statsmodel library is used to import the models that are solely statistics based. 

\begin{figure}[!t]
\centerline{\includegraphics[width=0.9\linewidth]{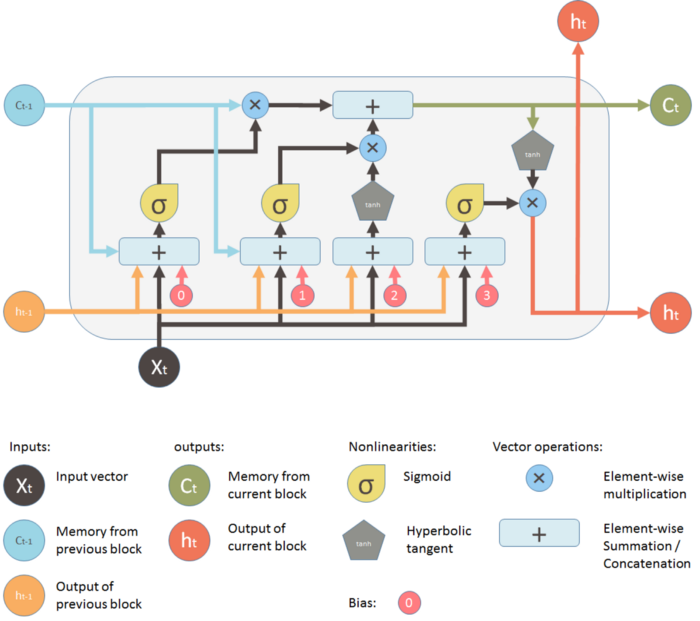}}
\caption{LSTM general architecture showing common components and their connectivity in the model \cite{yan2016understanding}}
\label{LSTM model picture}
\end{figure}

Finally, the LSTM is a modified recurrent neural network (RNN) with a chain like structure that is suitable for predictions where long-term dependency is an issue \cite{yan2016understanding}. The single tanh layer is substituted by a quad tanh layer as repeating module as shown in Fig. \ref{LSTM model picture}. Application of the model starts with identification of irrelevant data and subsequent removal of it. The sigmoid gate is used to carry out this operation. The inputs are the preceding (ht-1) input in t–1 (hidden), the current (Xt) input in t, and bias in bf \cite{bui2020power}. Two steps are taken to update the cell status based on the previous input Xt. This is the gate of the sigmoid function and the tanh. The sigmoid gate calculates old update information. The gate of Tanh scales the cell state of the new candidate value requirement. The resulting cell state is multiplied followed by addition to the old cell state to generate the current state of cell Ct.
Where cells are represented at time t-1 and t with Ct-1 and Ct, whereas b and W are respectively biases followed by cell-style matrices.
Finally, output values (ht) are calculated from the output cell state (Ct). The second step has similarity, but now there are updated information forming a new LSTM unit in the cell state.
The output gate's bias and weight patterns are represented by bo and Wo.

In Table \ref{LSTM Model Parameters}, the necessary parameters in the LSTM model is presented. The model uses sequential from tensorflow, the loss function is 'mean squared error' and the optimizer used is 'adam'. The number of epochs while training is 50 and the batch size is 70. The number of parameters shown in the table are all trainable parameters. 

\begin{table}[h]
\caption{LSTM Model showing Types of layers, shapes of output and parameter numbers}
\begin{center}
\begin{tabular}{|c|c|c|}
\hline
\textbf{Types of layers} & \textbf{Shapes of Output}& \textbf{Parameter Numbers}\\ \hline

LSTM               & (None, 100) & 50400 \\ \hline
Dropout               &           (None, 100) & 0   \\ \hline
Dense               &          (None, 1)      &  101      \\ \hline

\end{tabular}
\label{LSTM Model Parameters}
\end{center}
\end{table}

\begin{figure}[h]
\centerline{\includegraphics[width=0.8\linewidth]{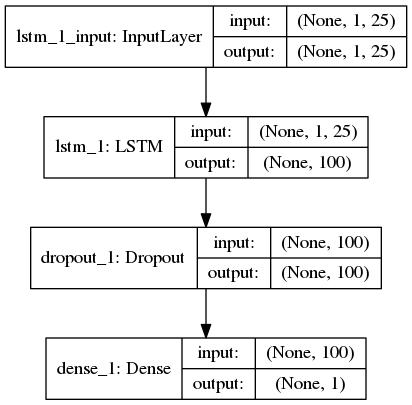}}
\caption{Plot of the model using keras.utils showing all the parameters and direction of flow}
\label{model plot}
\end{figure}

In Fig. \ref{model plot}, the model architecture has been presented with its hidden layers and shape of the inputs and outputs for each layers. It is comprehensible that even though we used neural net for forecasting time series data the model complexity is very low because of its usage of less hidden layers and fairly straightforward architectural build.

\begin{figure}[h]
\centerline{\includegraphics[width=0.8\linewidth]{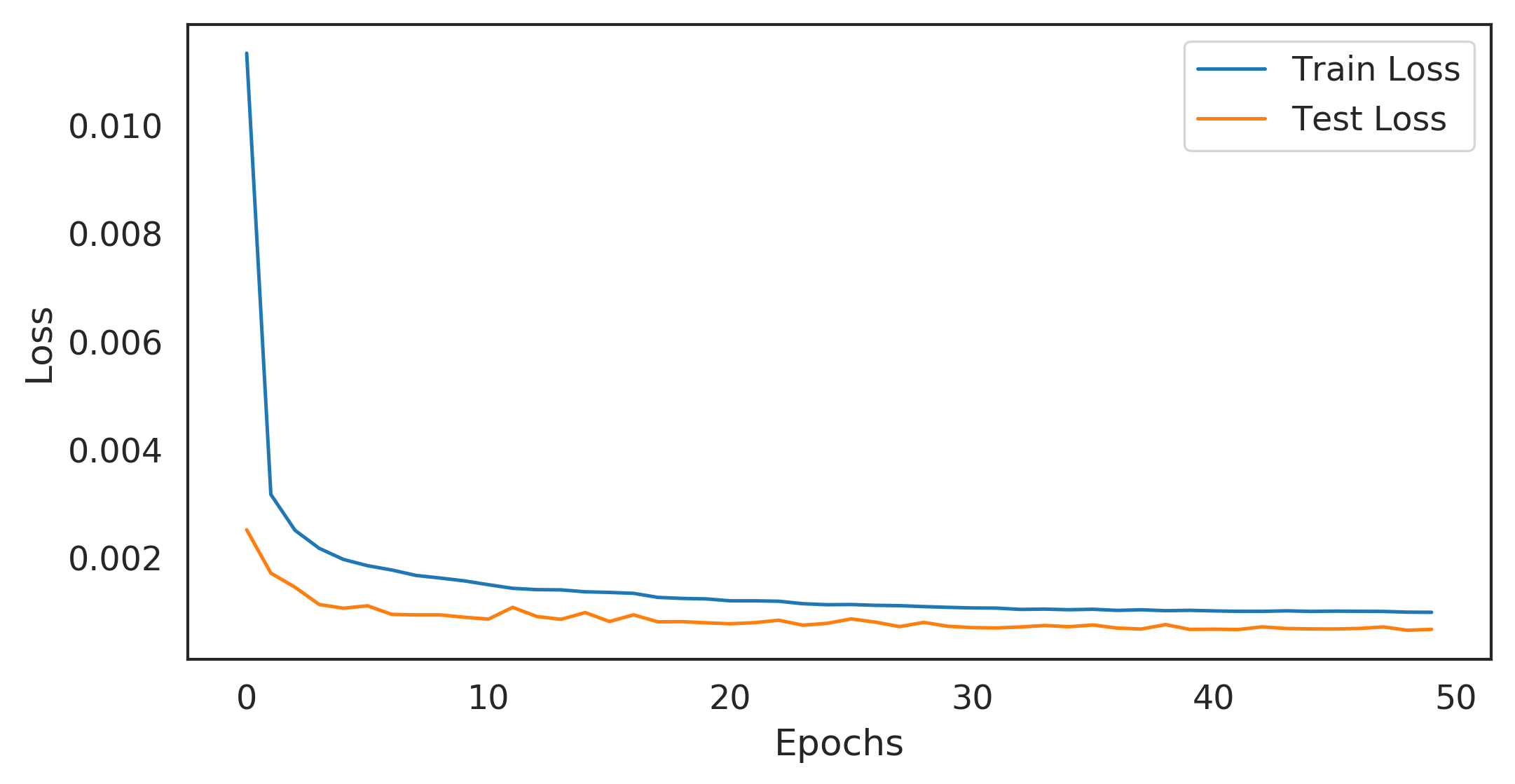}}
\caption{Loss curve of the model training showing the loss vs epoch for 50 epochs}
\label{loss curve}
\end{figure}

The Loss vs Epoch curve is shown in the Fig. \ref{loss curve} in which the progress of model while training is represented. Both training and testing loss decreases in a smooth fashion fairly quickly. This essentially means that the LSTM RNN model is optimized for the dataset that has been used in training and more computationally expensive models are not necessary in this case. 

The hardware spec in which the training and evaluation has been done is shown below:
\begin{itemize}
    \item GPU: 1xTesla K80 , compute 3.7, having 2496 CUDA cores, 12GB GDDR5 VRAM

    \item CPU: 1xsingle core hyper threaded Xeon Processors @2.3Ghz i.e (1 core, 2 threads)

    \item RAM: 12.6 GB Available

    \item Disk: 33 GB Available
\end{itemize}

\section{Performance Analysis}

\begin{figure}[h]
\centerline{\includegraphics[width=0.9\linewidth]{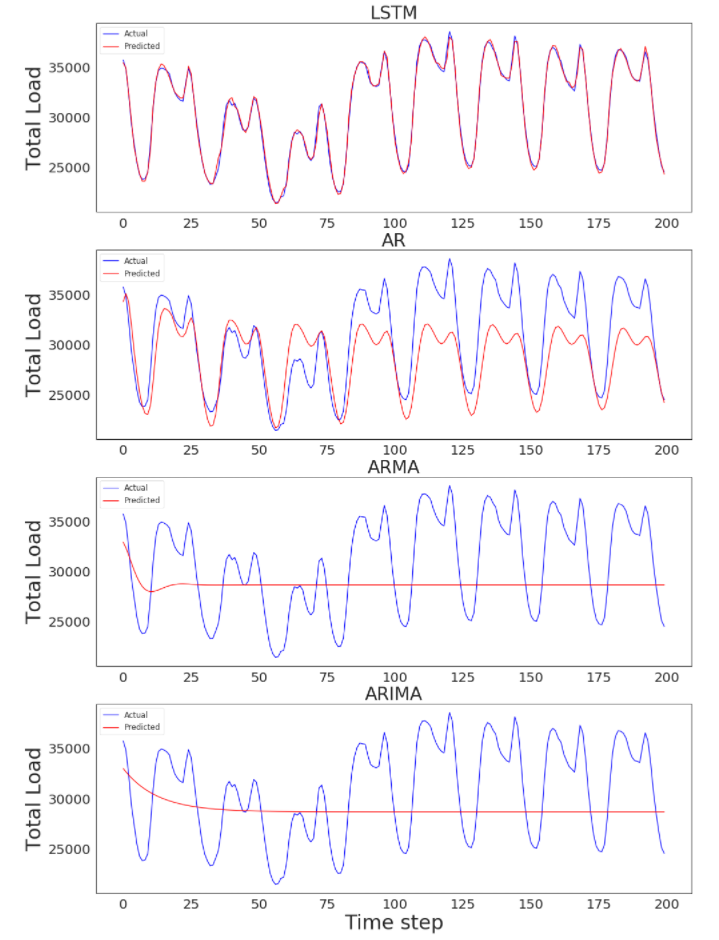}}
\caption{Visualizing prediction of different models}
\label{Visualizing presiction of different models}
\end{figure}

Several statistical models for example, Autoregressive Moving Average (ARMA), AutoRegressive (AR) and recurrent neural network model namely LSTM has been applied to the dataset. The actual and predicted outputs of all the above mentioned models have been shown in Fig. \ref{Visualizing presiction of different models}.

Mean absolute errors (MAE) are error measurements between pairing observations that show the same phenomenon. It is often used for the assessment of predictive model performance. For example, Y vs. X comprise comparisons of predicted versus observed, time vs. time, and a single measurement technique versus an alternative measurement technique, shown in Table \ref{Error rate}. MAE calculation is as such:

\begin{equation}
\operatorname{MAE}=\frac{\sum_{i=1}^{n}\left|y_{i}-x_{i}\right|}{n}
\end{equation}
Where
$x_{i}=$ true value,
$y_{i}=$ prediction value and
$n=$ no of total data points.

\begin{table}[h]
\caption{Error rate comparison of different models containing Mean Absolute Error and Mean Absolute Percentile Error}
\begin{center}
\begin{tabular}{|c|c|c|}
\hline
\textbf{Model Name} & \textbf{MAE (MWh)}& \textbf{MAPE}\\ \hline

LSTM               &            361.406 & 1.22\\ \hline
AR               &           3729.264 & 13.33\\ \hline
MA               &           34497.7 & 59.13\\ \hline
ARMA      &          3862.929 & 13.81\\ \hline
ARIMA                      &          3865.296 & 35.88\\ \hline

\end{tabular}
\label{Error rate}
\end{center}
\end{table}

From the prediction output in Fig. \ref{Visualizing presiction of different models} and the Mean Absolute Error values from Table \ref{Error rate}, it is obvious that the LSTM model perform much better in forecasting power demand than the other models. The non-stationary nature of the data is the reason that the neural net models provide better result than statistical models. Furthermore, the Mean Absolute Percentage Error (MAPE) shows favorable results in case of LSTM, resulting in highly accurate forecasting result compared to other methods.

\section{Conclusion}
This paper shows that power demand forecasting using LSTM has higher accuracy and lower error rate than traditional statistics based forecasting models such as AR, ARMA and ARIMA. So, application of LSTM with further hypertuned parameters and optimizations can be an effective tool for power demand forecasting and yield better results. Also, the high prediction accuracy could also be an indicator that other RNN based models such as Gated Recurrent Unit might also be suitable as much as or even more than LSTM. The efficiency of this model will depend heavily on the availability of the long term past data. So, further research into effectiveness of RNNs in demand forecasting could demonstrate higher accuracy.

\bibliographystyle{./bibliography/IEEEtran}
\bibliography{./bibliography/IEEEabrv,./bibliography/IEEEexample}

\end{document}